# Individual Causal Inference with Structural Causal Model

Daniel T. Chang (张遵)

*IBM (Retired)* dtchang43@gmail.com

**Abstract:**. Individual causal inference (ICI) uses causal inference methods to understand and predict the effects of interventions on individuals, considering their specific characteristics / facts. It aims to estimate individual causal effect (ICE), which varies across individuals. Estimating ICE can be challenging due to the limited data available for individuals, and the fact that most causal inference methods are population-based. Structural Causal Model (SCM) is fundamentally population-based. Therefore, causal discovery (structural learning and parameter learning), association queries and intervention queries are all naturally population-based. However, exogenous variables (U) in SCM can encode individual variations and thus provide the mechanism for individualized population per specific individual characteristics / facts. Based on this, we propose ICI with SCM as a "rung 3" causal inference, because it involves "imagining" what would be the causal effect of a hypothetical intervention on an individual, given the individual's observed characteristics / facts. Specifically, we propose the indiv-operator, indiv(W), to formalize/represent the population individualization process, and the individual causal query, P(Y | indiv(W), do(X), Z), to formalize/represent ICI. We show and argue that ICI with SCM is inference on individual alternatives (possible), not individual counterfactuals (non-actual).

## 1 Introduction

*Individual causal inference (ICI)* [9-12], a.k.a. *individualized causal inference*, uses causal inference methods to understand and predict the effects of interventions on individuals, considering their specific characteristics / facts. Instead of estimating average causal effects across a population, ICI aims to estimate *individual causal effect (ICE)*, which varies across individuals. It is particularly relevant in *personalized medicine, personalized decision-making*, and other fields where interventions are tailored to individual needs, and has the potential to revolutionize these fields by enabling more targeted and effective interventions.

Estimating ICE can be challenging due to the limited data available for individuals. Furthermore, most causal inference methods are population-based. It is important to note that ICE is not *ITE (individual treatment effect)* [6-8]. ITE is defined for a counterfactual scenario: treatment vs. no treatment, with the latter being *non-actual*. ITE cannot be observed or computed. ICE, on the other hand, deals with alternative scenarios which are *possible*. ICE can be observed, in principle, and computed.

*Structural Causal Model (SCM)* [1-5] is fundamentally population-based. Therefore, causal discovery (structural learning and parameter learning), association queries and intervention queries are all naturally population-based. However, *exogenous variables (U)* in SCM can encode individual variations and thus provide the mechanism for *individualized population* per specific individual characteristics / facts. The population is only partially individualized because the individual characteristics / facts observed are usually limited (*partial observability*).

In this paper, we propose *ICI with SCM* as a *"rung 3" causal inference,* because it involves "imagining" what would be the causal effect of a hypothetical intervention on an individual, given the individual's observed characteristics / facts. ICI with SCM is necessarily population-based. However, importantly, the population is individualized per specific individual characteristics / facts. Specifically, we propose the *indiv-operator, indiv(W),* to formalize/represent the population individualization process, and the *individual causal query, $P(Y \mid indiv(W), do(X), Z)$* to formalize/represent ICI.

This paper is organized as follows. For background information and completeness of presentation and discussion, we first discuss SCM, association queries and intervention queries. We then discuss ICI with SCM, focusing on individual causal queries, $P(Y \mid indiv(W), do(X), Z)$, and the indiv-operator, indiv(W). Finally, we show and argue that ICI with SCM is inference on individual alternatives (possible), not individual counterfactuals (non-actual).

## 2 Structural Causal Model (SCM)

*Structural Causal Model (SCM)* [1-5] is typically defined as a tuple *M=(V, U, F)*, where:

- *V (Endogenous Variables):* This is a set of variables $V_i$ whose values are determined by factors *within* the model. These are the variables whose *causal mechanisms* we are trying to understand. They can be *discrete or continuous*, and they can be *observed or latent (unobserved)*. For convenience in later discussions, we designate special-role endogenous variables as:
    - *Y*: query variables
    - *Z*: evidence variables
    - *X*: intervention variables
    - *W*: individual characteristic / fact variables
- *U (Exogenous Variables)*: This is a set of variables $U_i$ which represent factors *outside* the model that influence the system. They are *unobserved*. Each $U_i$ is associated with a *probability distribution $P(U_i)$*. There is *one $U_i$ per $V_i$*.

- *F (Structural Equations)*: This is a set of structural equations, one for each endogenous variable $V_i$. Each equation defines the value of $V_i$ as a function of its direct causes (parents) and its corresponding exogenous variable $U_i$. Formally, for each $V_i$, there is a *deterministic function $f_i$*:

  $$V_i = f_i(Pa(V_i), U_i)$$

  where $Pa(V_i)$ denotes the set of direct causes (parents) of $V_i$ within the model.

The key implications and components of SCM are:

- *Determinism:* Each structural equation is deterministic. The probabilistic nature of SCM comes solely from the probability distributions of the exogenous variables $P(U_i)$.
- *Causal Graph***:** An SCM inherently implies a causal graph, a *directed acyclic graph (DAG)*, where nodes represent *endogenous variables V* and directed edges represent *direct (parent-child) causal relationships* between them. The absence of an arrow between two endogenous variables is a strong assumption, implying no direct causal effect between them.
- *Conditional probabilities*: An SCM inherently implies the conditional probability of a variable given its parents $P(V_i | Pa(V_i))$, where the probability of the variable $V_i$ when the parents are set to some values by intervention is the same as it is when the parents are observed to have those values.
- *Acyclicity:* There are no directed cycles in the causal graph. This property is crucial for ensuring that the *joint probability distribution of V* can be uniquely defined and factorized:

  $$P(V_1, V_2, \ldots, V_n) = \prod_{i=1}^{n} P(V_i | Pa(V_i))$$

  This factorization is based on the *local Markov property*, which states that each variable $V_i$ in the causal graph is conditionally independent of its non-descendants given its parents.

- *Population-based*: SCM is fundamentally population-based. This is reflected in the fact that causal discovery (structural learning and parameter learning), association queries and intervention queries are all naturally population-based. The population characteristics are encoded in the *exogenous variables U*, which nonetheless allow for/can encode individual variations.

## 2.1 Exogenous Variables U

The exogenous variables U are the *hidden parameters* of SCM. They are:

- *Unobserved Factors*: They represent all the unobserved factors that influence the respective endogenous variables V but are not explicitly part of the model.
- *Source of Variability*: They are the source of the inherent variability or randomness in the system. While the structural equations themselves are deterministic functions, the exogenous variables are random variables with a certain probability distribution $P(U_i)$.

The *values/distributions* of exogenous variables U in an SCM are obtained primarily through:

- Assuming specific probability distributions for *a particular population*: This is common in theoretical analysis, simulations, or when building a model (e.g., assuming they are Gaussian distributions).
- Estimating the values/distributions from *observed data of a particular population*: If we have data for the endogenous variables V, we can infer the values/distributions of the exogenous variables U that best fit the data according to the SCM's structural equations.

The exogenous variables (U) play a *crucial role* in SCM enabling association queries, intervention queries and individual causal queries, as can be seen from subsequent discussions.

## 3 Association Queries: "Seeing", Population Based

Association queries ask *"What is?"* questions. With SCM, an *association query P(Y | Z)* refers to a *conditional probability query*. To calculate P(Y | Z), where Y is a *query variable* (or set of variables) and Z is an *evidence variable* (or set of variables), we generally follow these steps:

1. Start with the joint probability distribution:

    $P(V_1, V_2, \ldots, V_n) = \prod_{i=1}^{n} P(V_i | Pa(V_i))$

2. Formulate the joint probability of query and evidence:

    We want P(Y | Z). According to the rule of conditional probability,

$$P(Y \mid Z) = P(Y, Z) / P(Z).$$

So, the first step is to compute P(Y, Z). This involves *marginalizing (summing out)* all variables that are not Y or Z from the full joint probability distribution. Let H be the set of non-query variables (all variables in the causal graph that are not Y and not Z).

$$P(Y, Z) = \sum_h P(Y, Z, H=h),$$

where h ε states(H) and P(Y, Z, H=h) is the probability of a specific instantiation of Y, Z and H, which can be computed using the joint probability factorization from step 1.

3. Calculate the marginal probability of the evidence:

Next, we need P(Z), which acts as a normalization constant. This is obtained by summing out the query variable(s) Y from the joint probability of Y and Z:

$$P(Z) = \sum_y P(Y=y, Z),$$

where y ε states(Y) and P(Y=y, Z) is the probability of a specific instantiation of Y and Z.

4. Normalize to get the conditional probability:

Finally, we divide the joint probability of query and evidence by the marginal probability of the evidence:

$$P(Y \mid Z) = P(Y, Z) / P(Z)$$

Performing the direct summation in the above steps for large causal graphs can be computationally intractable (NP-hard in general). As such, standard *probabilistic inference algorithms* are to be used to compute P(Y | Z) efficiently:

- *Variable Elimination*: This algorithm systematically eliminates non-query variables by pushing summations into the product, exploiting the causal graph's conditional independencies to avoid forming the full joint distribution. It effectively reorders the summations and multiplications.

- *Belief Propagation*: This algorithm transforms the causal graph into a tree of cliques (maximal complete subgraphs), allowing for message passing between cliques to efficiently compute marginal and conditional probabilities for all variables.
- *Approximate Inference*: For very large or complex causal graphs where exact inference is too slow, approximate methods like *Monte Carlo Markov Chain (MCMC) sampling* or *variational inference* can be used to estimate the probabilities.

### 3.1 Exogenous Variables U

With SCM, the conditional probabilities, $P(V_i | Pa(V_i))$, can be calculated from the probability distributions, $P(U_i)$, of exogenous variables U. The following is a simple example for illustration.

SCM:

1. $Z = U_Z$
2. $X = Z + U_X$
3. $Y = X + Z + U_Y$

Conditional Probabilities (Assuming $U_Z$, $U_X$, $U_Y$ are mutually independent):

1. $P(Z=z)$: Derived directly from $P(U_Z)$.
    - $P(Z=z) = P(U_Z=z)$
2. $P(X=x | Z=z)$: Derived from $P(U_X)$.
    - $P(X=x | Z=z) = P(U_X=x−z)$
3. $P(Y=y | X=x, Z=z)$: Derived from $P(U_Y)$.
    - $P(Y=y | X=x, Z=z) = P(U_Y=y−x−z)$

## 4 Intervention Queries: "Doing", Population Based

Intervention queries ask *"What if?"* questions. With SCM, an *intervention query $P(Y | do(X), Z)$* investigates a *causal effect*. It simulates an experiment where we directly manipulate the *intervention variable (or set of variables) X*, thereby overriding any causal influences on X, and observe the subsequent effect on Y.

The key to computing P(Y | do(X), Z) is a conceptual or actual modification of the original causal graph. This modification is often called *graph surgery*. When we intervene and set X to a specific value x (i.e., do(X=x)), we are essentially removing all incoming edges to X in the causal graph. The value of X is no longer determined by its usual causes; it's now determined by our *external intervention*.

The formal procedure for calculating P(Y | do(X=x), Z) is as follows:

1. Modify the Graph:

    a) Take the original causal graph G.
    b) Create a new graph, $G_{do(X=x)}$, by removing all edges that point into X.
    c) This reflects X's value is no longer a probabilistic outcome of its parents; it's fixed by the intervention to x.

2. Formulate the Truncated Factorization (or Manipulated Probability Distribution):

    a) The joint probability distribution of the variables in this new graph $G_{do(X=x)}$ is given by the *truncated factorization*:

    $$P_{do(X=x)}(V_1, V_2, \ldots, V_n) = \prod_{V_i \in V \setminus X} P(V_i | Pa(V_i)) \, I(X=x),$$

    where I(X=x) is an *indicator function* that is 1 if the variables in X take the value x, and 0 otherwise.

    b) Essentially, the term P(X | Pa(X)) is removed from the joint probability product, and X is simply fixed to x.

3. Perform Standard Inference on the Modified Distribution:

    Once we have this modified joint probability distribution, $P_{do(X=x)}(V_1, V_2, \ldots, V_n)$, we can calculate P(Y | do(X=x), Z) by performing standard marginalization, just like in a regular causal graph:

    $$P_{do(X=x)}(Y, Z) = \sum_h P_{do(X=x)}(Y, Z, H=h),$$

    where H is the set of variables in the causal graph that are not Y and not Z. It follows then:

    $$P_{do(X=x)}(Y | Z) = P_{do(X=x)}(Y, Z) / P_{do(X=x)}(Z).$$

## 4.1 Exogenous Variables U

The exogenous variables (U) enable intervention queries with SCM. The following simple example illustrates the calculation of causal effects in an intervention scenario (do(X=0), Y = ?). To derive Y under the intervention do(X=0) with a given SCM, we need to substitute the intervention directly into its structural equations:

SCM:

1. $Z = U_Z$
2. $X = Z + U_X$
3. $Y = X + Z + U_Y$

Step 1: Intervention (do(X=0)):

Modified SCM ($M_{do(X=0)}$):

1. $Z = U_Z$
2. $X = 0$
3. $Y = X + Z + U_Y$

Step 2: Inference (Calculate Vi):

- Inputs:
    - Modified SCM: $X = 0$, $Z = U_Z$, $Y = X + Z + U_Y$
- Calculation:
    - $X = 0$
    - $Z = U_Z$
    - $Y = X + Z + U_Y = U_Z + U_Y$
- Result: $Y = U_Z + U_Y$ (The outcome of intervention do(X=0) is *predicated* on the exogenous variables U.)

The conditional probabilities, $P(V_i \mid Pa(V_i))$, can be calculated from the probability distributions, $P(U_i)$, of exogenous variables U. Assuming $U_Z$, $U_X$, $U_Y$ are mutually independent:

1. $P(Z=z)$: Derived directly from $P(U_Z)$.

- $P(Z=z) = P(U_Z=z)$
2. $P(X)$: $X=0$
   - $P(X=x) = I(X=0)$
3. $P(Y=y \mid X=x, Z=z)$: Derived from $P(U_Y)$.
   - $P(Y=y \mid X=0, Z=z) = P(U_Y=y-z)$

## 5 Individual Causal Queries: "Imagining", Individualized-population Based

*Individual causal inference (ICI)* [9-12], a.k.a. *individualized causal inference*, uses causal inference methods to understand and predict the effects of interventions on individuals, considering their specific characteristics / facts. Instead of estimating average causal effects across a population, ICI aims to estimate *individual causal effect (ICE)*, which varies across individuals. Estimating ICE, however, can be challenging due to the limited data available for individuals. Furthermore, most causal inference methods are population-based.

*ICI with SCM* is a *"rung 3" causal inference,* because it involves "imagining" what would be the causal effect of a hypothetical intervention on an individual, given the individual's observed characteristics / facts. ICI with SCM is necessarily population-based. However, importantly, the population is *individualized* per specific individual characteristics / facts. Specifically, ICI with SCM consists of the *indiv-operator, indiv(W)* which formalizes/represents the population individualization process, and the *individual causal query, $P(Y \mid indiv(W), do(X), Z)$* which formalizes/represents ICI.

Individual causal queries ask *"What if I had acted differently?"* questions. The key is that they deal with individuals ("I"), observed facts ("acted"), and hypothetical actions / interventions ("had acted differently")..With SCM, an individual causal query $P(Y \mid indiv(W), do(X), Z)$, i.e., ICI, involves a *three-step process*:

1. *Abduction (indiv(W))*: This is the *crucial* first step where we use the observed *individual characteristics / facts (W)* to infer the specific values of the exogenous variables (U) that produce the characteristics / facts.
2. *Intervention (do(X))*: In this step, we modify the original SCM (M) to create a new SCM ($M_{do(X)}$) that represents the hypothetical intervention scenario (with variables X).
3. *Inference ($P_{ICI}(Y \mid Z)$)*: In this final step, we use the modified SCM ($M_{do(X)}$) and the specific values of the exogenous variables (U) that are abducted in Step 1 to infer the value of the outcome variables (Y) under the hypothetical intervention scenario.

The abduction step, *indiv(W)*, is the *crucial step* for ICI. It *individualizes the population* (whose characteristics are encoded in the exogenous variables U) per the observed individual characteristics / facts (W). Abduction here refers to inferring the values of the exogenous variables (U) that are based on/consistent with the individual characteristics / facts (W). These inferred $U_i$ values are then used to evaluate individual causal queries.

For convenience of discussion, we assume an SCM with 50 $V_i$ / $U_i$ and individual factual observations (X=1, Y=1). There will generally be *infinitely many* combinations of $U_i$ values that could produce the observed (X=1, Y=1). This is because the system is underdetermined; we have 50 unknown variables $U_i$, but only two observed variables (X and Y).

Given the large number of unknown variables $U_i$ (50) and no assumptions on functional form, a direct, universally applicable analytical solution for abducing $U_i$ is generally intractable:

- *Deterministic Solution*: Not feasible with partial observations (X=1, Y=1) without strong additional assumptions or full observability. There's no unique set of $U_i$ to abduce deterministically.
- *Probabilistic Solution*: This is the most *appropriate and robust* approach.
  - Assume *prior distributions* for all $U_i$.
  - Use Markov Chain Monte Carlo (MCMC) methods to sample from the *posterior distributions* of $U_i$ given the individual factual observations (X=1, Y=1). The posterior distributions represent/encode the *individualized population*.

We discuss the details of each solution in the following.

### 5.1 indiv(W): Deterministic Abduction of Exogenous Variables U

In an SCM, if we know the precise functional forms $f_i$ and the factual values for all $V_i$, then we can in principle uniquely determine the values of $U_i$:

- *Ideal Scenario (Full Observability)*: If we observe the factual values for all Vi, and if all $f_i$ are *invertible* with respect to $U_i$ (given $Pa(V_i)$), then we can directly calculate:

$$U_i^{fact} = f_i^{-1}(V_i^{fact}, Pa(V_i)^{fact})$$

This will give us a unique set of $U_i^{fact}$ values consistent with the facts. (A simple example is shown later when we discuss inference on individual alternatives vs. individual counterfactuals.)

- *Realistic Scenario (Partial Observability)*: With only (X=1, Y=1) observed and 50 $U_i$ variables, we have a severely underdetermined system.
  - Implication for Abduction: There will be infinitely many combinations of $U_i$ values that can lead to X=1 and Y=1.
  - No Unique Deterministic Solution: We cannot uniquely identify $U_i$ values from partial observations. We will need to make strong assumptions, such as specific functional forms or additional observations, to narrow down the possibilities.
  - Path-specific Abduction: If we consider specific paths in the causal graph, and if the values along those paths were observed, we might be able to constrain some $U_i$. However, with just (X=1, Y=1), this is still generally insufficient, as shown below.

## "Update" Solution

For *path-specific abduction*, an "update" solution for $U_i$ based on partial observations might be possible. To do so, we need to introduce some form of flexibility or uncertainty into the model, even if the underlying mechanisms are deterministic. For *ICI*, we can take advantage of the following:

- The 3-step process implies that after observing the individual factuals, we fix the $U_i$ values to those that would have generated the individual factual observations.
- If our original "known" $U_i$ didn't produce (X=1, Y=1), then we need to find the "closest" set of $U_i$ that does produce the facts.
- This involves finding the $U_i^*$ such that $V^{factual} = SCM(U_i^*)$ and $U_i^*$ is "closest" to some baseline or prior beliefs about $U_i$. This is a form of *constrained optimization*.

Given (X=1, Y=1), if we have a mechanism to deterministically infer $U_i$ (e.g., by inverting the functions, which requires unique invertibility), we could proceed with *constrained optimization* as follows:

- For any observed variable $V_k$, if its structural equation $f_k$ is invertible with respect to $U_k$ (given $Pa(V_k)$), then $U_k = f_k^{-1}(V_k, Pa(V_k))$.
- The "update" would involve:

  - Directly affected $U_i$: $U_X$ and $U_Y$.
  - Indirectly affected $U_i$: Any $U_i$ whose corresponding $V_i$ values are implied or constrained by the observations (X=1, Y=1) and the causal graph. For example, if $X \leftarrow Z$ and observing X uniquely determines Z, then $U_Z$ would be constrained or deterrmined. This requires working backward through the causal graph.
  - Unconstrained $U_i$: For many $V_i$ that are not ancestors of X or Y, or whose effects are entirely blocked by observed variables, their $U_i$ would remain unconstrained by the observations (X=1, Y=1). Their "values" would remain as their initial "known" values or default to some baseline.

## 5.2 indiv(W): Probabilistic Abduction of Exogenous Variables U

With no assumptions on functional form and partial observations, a *probabilistic approach* is far more appropriate and robust for abducing $U_i$. We treat $U_i$ as random variables with known (or assumed) distributions.

1. *Define Distributions for $U_i$*: We assume *prior distributions* for each $U_i$, e.g., $U_i \sim N(0, \sigma_i^2)$ or $U_i \sim Uniform(a,b)$. These distributions capture the variability and uncertainty about the true values of the exogenous variables.
2. *Bayesian Inference for Abduction*: The abduction of $U_i$ becomes a problem of Bayesian inference: we want to find the posterior distribution of $U_i$ given the individual factual observations (X=1, Y=1).

   $P(U_1,\ldots,U_n \mid X=1, Y=1) = P(X=1, Y=1 \mid U_1,\ldots,U_n)\, P(U_1,\ldots,U_n) \,/\, P(X=1, Y=1)$

   - Likelihood $P(X=1, Y=1 \mid U_1,\ldots,U_n)$: This term represents the probability of observing (X=1, Y=1) given specific values for all $U_i$. In an SCM where $V_i$ are functions of $Pa(V_i)$ and $U_i$, this likelihood would typically be 0 or 1. If the given $U_i$ values, through the SCM, produce X=1 and Y=1, the likelihood is 1; otherwise, it's 0.
   - Monte Carlo Methods: Since analytical solutions are impossible except for certain functional forms, we would resort to Monte Carlo methods, specifically Markov Chain Monte Carlo (MCMC) techniques.
     - Sampling from the Posterior: MCMC allows us to draw samples from the posterior distribution $P(U_1,\ldots,U_n \mid X=1, Y=1)$. Each sample $U_i^{(k)}$ represents a plausible set of exogenous values consistent with the facts.

- Algorithm Outline:

    1. Initialize Ui values.
    2. Propose new $U_i$ values (e.g., by sampling from a proposal distribution).
    3. Run the SCM forward with the proposed $U_i$ values to see what values of X and Y are generated.
    4. Accept or reject the proposed $U_i$ values based on their consistency with (X=1, Y=1) and the prior probabilities, using the MCMC acceptance criteria.
    5. Repeat many times to build up a sample of $U_i$ values.

Note that P(X=1, Y=1) is the normalizing constant for the posterior distribution of Ui. It is obtained by integrating the product of the likelihood and the prior over the entire space of Ui. While its direct calculation is often intractable for complex SCMs, it is implicitly handled by MCMC sampling algorithms. This is one of the major advantages of MCMC.

### Intervention (do(X)) and Inference ($P_{ICI}(Y | Z)$)

Once we have a collection of abduced sample sets $U_i^{(k)}$ from the posterior distributions, we can proceed wih the *intervention (do(X)) step*. For each sampled set $U_i^{(k)}$:

1. Set the intervention: do(X=x). This means we override the causal mechanism for X and set its value to x.
2. Propagate the values through the SCM: Using the abduced $U_i^{(k)}$ and the modified SCM (with X set to x), compute the values of all other variables, including Y.
3. Record the resulting Y value, $Y^{(k)}$.

The distribution of these $Y^{(k)}$ values provides an estimate of the *inference $P_{ICI}(Y | Z)$*.

## 6 ICI with SCM: Inference on Individual Alternatives, Not Individual Counterfactuals

We show and argue that ICI with SCM is inference on *individual alternatives*, not individual counterfactuals. The following is a concrete example to backup our argument. In the example, the intervention variable X can take the value of 1 or 0, and the individual characteristic / fact variables (W) are: (X=1, Y=10, Z=2) .

SCM:

1. $Z = U_Z$

2. $X = Z + U_X$

3. $Y = X + Z + U_Y$

*Factual observation for an individual: Xobs = 1, Yobs = 10, Zobs = 2.*

Step 1: Abduction (Resolve $U_i$ from the observed individual facts)

- From Zobs = 2 $\Rightarrow U_Z^* = 2$

- From Xobs = 1 and Zobs = 2: $1 = 2 + U_X \Rightarrow U_X^* = -1$

- From Yobs = 10, Xobs = 1, Zobs = 2: $10 = 1 + 2 + U_Y \Rightarrow U_Y^* = 7$

- The resolved $U_i^*$ represent the specific unobserved characteristics / facts of this individual, who possesses the observed facts.

Step 2: Hypothetical Intervention

Hypothetical Question: "What would this individual's Y be if she had taken the action X=0?" (do(X=0))

Modified SCM ($M_{do(X=0)}$):

1. $Z = U_Z$

2. $X = 0$

3. $Y = X + Z + U_Y$

Step 3: Inference (Calculate hypothetical $V_i$)

- Inputs:
    - Resolved $U_i^*$: $U_Z^* = 2$, $U_X^* = -1$, $U_Y^* = 7$
    - Modified SCM: $X = 0$, $Z = U_Z$, $Y = X + Z + U_Y$
- Calculation:
    - $Z_{hypo} = U_Z^* = 2$
    - $X_{hypo} = 0$
    - $Y_{hypo} = X_{hypo} + Z_{hypo} + U_Y^* = 0 + 2 + 7 = 9$

- Result: For this specific individual, who took the action (X=1) and had an outcome of Y=10, her outcome would be Y=9 if she had taken the action (X=0).

This result allows us to say something like: "For this person, the hypothetical action would cause an increase in Y of 10 – 9 = 1 unit." This is an *alternative* statement, attributing a causal effect to the hypothetical action for a specific individual, based on exogenous variables U abducted from observed individual facts.

It is important to note that we can *reproduce the individual observed facts* if we use the factual intervention (X=1) in the calculation:

- Calculation:
    - $Z_{fact} = U_Z^* = 2$
    - $X_{fact} = 1$
    - $Y_{fact} = X_{fact} + Z_{fact} + U_Y^* = 1 + 2 + 7 = 10$

This shouldn't be surprising since the resolved $U_i^*$ are derived from the observed individual facts. However, this highlights that the *hypothetical intervention (X=0)* takes place in the *observed world (X=0)*, not a hypothetical world. Therefore, the hypothetical intervention (X=0) is an *alternative*. In fact, if X were multivalued (0, 1, 2, 3, …), we could try other alternative hypothetical interventions (X=2, X=3, …) and investigate their effects.

The major characteristics of *alternatives* are:

- *Different-from-fact*: They represent something that did not happen.
- *Possible*: They represent something that could have happened, or can happen in the future.
- *Feasible*: They are within the realm of what is achievable...

## Individual Counterfactuals

In contrast to alternatives, the major characteristics of *counterfactuals* are:

- *Contrary-to-fact*: They represent something that did not happen.
- *Non-actual*: They are not part of the current reality. They did not happen and they could not have happened.
- *Hypothetical*: They exist purely in the realm of "what if".

Recall that the abducted exogenous variables $U_i^*$ are derived from observed individual facts. Therefore, they represent an *"actual" observed world*. They do not represent a "non-actual" hypothetical world.

For SCM to be able to represent a *"non-actual" hypothetical world*, it would need to be able to retract the factual observation for an individual, either by going back "in time" or removing what has happened, thus enabling individual counterfactual interventions. *SCM has no mechanism to represent a "non-actual" hypothetical world.*

Counterfactuals are *non-actual*, meaning they *did not and could not have occurred* in reality. They are *mutually exclusive* with the observed event and *impossible* in the observed world. For counterfactuals, asking "What if I had done differently?" is necessary, but not sufficient. For example, the question "What if the student had taken Test B?" is an alternative question, not a counterfactual question. A counterfactual question would be "The student took Test A, preventing from ever taking Test B. What if the student had taken Test B?" The "The student took Test A, preventing from ever taking Test B." part is essential, representing *mutual exclusivity and impossibility*. Although this part usually is represented implicitly by context, it must be there for the question to be counterfactual. *SCM has no mechanism to represent mutual exclusivity and impossibility.*

In summary, we argue that:

- Inference on individual alternatives involves *hypothetical interventions* in an *"actual" observed world*, whereas
- Inference on individual counterfactuals involves *hypothetical interventions* in a *"non-actual" hypothetical world*,

Therefore, ICI with SCM is inference on *individual alternatives (possible), not individual counterfactuals (non-actual)*.

## 7 Conclusion

Individual causal inference (ICI) uses causal inference methods to understand and predict the effects of interventions on individuals, considering their specific characteristics / facts. It is particularly relevant in personalized medicine, personalized decision-making, and other fields where interventions are tailored to individual needs, and has the potential to revolutionize these fields by enabling more targeted and effective interventions.

In this paper we propose ICI with Structural Causal Model (SCM). Our major contributions are:

- We note that SCM is fundamentally population-based. Therefore, causal discovery (structural learning and parameter learning), association queries and intervention queries are all naturally population-based. However, exogenous variables (U) can provide the mechanism for individualized population per specific individual characteristics / facts. They crucially allow for/can encode individual variations.

- We propose **ICI with SCM** as a "rung 3" causal inference, because it involves "imagining" what would be the causal effect of a hypothetical intervention on an individual, given the individual's observed characteristics / facts..

    - ICI with SCM is necessarily population-based. However, importantly, the population is individualized per specific individual characteristics / facts. That is, the joint probability distribution of endogenous variables (V) is adjusted to be based on/consistent with the specific individual characteristics / facts (W).
    - **Individualized population** is achieved by abducting U given W.

- We propose the **indiv-operator, indiv(W)**, to formalize/represent the abduction process.

- We propose the **individual causal query, P(Y | indiv(W), do(X), Z)** to formalize/represent ICI with SCM.

- We show and argue that ICI with SCM is **inference on individual alternatives (possible)**, not individual counterfactuals (non-actual).

**Acknowledgement:** Thanks to my wife Hedy (郑期芳) for her support.

# Appendix. Alternative vs. Counterfactual Reasoning

While often used interchangeably or in closely related contexts, "alternative reasoning" and "counterfactual reasoning" have distinct nuances.

*Alternative reasoning* is a broader term that encompasses any cognitive process involving the consideration of different possibilities, options, or courses of action. It's about exploring what *could be* or *could have been* in a general sense, without necessarily implying that the imagined scenario is contrary to a known fact. Key aspects of alternative reasoning are:

- Focus: Exploring various possibilities, options, or solutions.
- Scope: Can apply to past, present, or future situations.
- Nature: Can be about generating new ideas, evaluating different strategies, or simply acknowledging that there are multiple ways things could unfold.
- Example: "What alternative routes could I have taken to avoid traffic?"

*Counterfactual reasoning* is a specific type of alternative reasoning. It involves thinking about what *would have happened* if only a past event had been different, i.e., "*contrary to fact (could not be* and *could not have been)*." It explicitly deals with scenarios that did *not* occur but are imagined as if they had. This often involves a *"if only"* structure. Key aspects of counterfactual reasoning are:

- Focus: Imagining a past event as having been different and then exploring the consequences of that hypothetical change.
- Scope: Primarily deals with past events, though it can inform future decisions by learning from hypothetical pasts.
- Nature: Often used for:
    - Causal inference: Understanding cause-and-effect relationships ("If only I hadn't stayed up so late, I wouldn't have been so tired for the exam.").
    - Learning and decision-making: Reflecting on past mistakes or successes to inform future behavior ("If only I had studied harder, I would have passed.").

For comparison, the major characteristics of *alternatives* are:

- *Different-from-fact:* They represent something that did not happen.
- *Possible:* They represent something that could have happened, or can happen in the future.
- *Feasible:* They are within the realm of what is achievable.

In contrast, the major characteristics of *counterfactuals* are:

- *Contrary-to-fact:* They represent something that did not happen.
- *Non-actual:* They are not part of the current reality. They did not happen and they could not have happened.
- *Imaginary:* They exist purely in the realm of "if only".

Both alternative reasoning and counterfactual reasoning engage in *"what if I had"* scenarios. *Alternative reasoning*, in its essence of "considering different options or choices", is inherently *prospective*, focusing on actions and decisions—e.g., "*if I had done X.*" Conversely, *counterfactual reasoning* is explicitly *retrospective*, dealing with "imaginary alternatives to past events" through expressions like "*if only I had done X.*"

All counterfactual reasoning is a form of alternative reasoning, but not all alternative reasoning is counterfactual.